\documentclass[sigconf]{acmart}

\AtBeginDocument{%
  \providecommand\BibTeX{{%
    \normalfont B\kern-0.5em{\scshape i\kern-0.25em b}\kern-0.8em\TeX}}}

\copyrightyear{2025}
\acmYear{2025}
\setcopyright{rightsretained}
\acmConference[KDD '25]{Proceedings of the 31st ACM SIGKDD Conference on Knowledge Discovery and Data Mining V.1}{August 3--7, 2025}{Toronto, ON, Canada}
\acmBooktitle{Proceedings of the 31st ACM SIGKDD Conference on Knowledge Discovery and Data Mining V.1 (KDD '25), August 3--7, 2025, Toronto, ON, Canada}
\acmDOI{10.1145/3690624.3709254}
\acmISBN{979-8-4007-1245-6/25/08}

\usepackage{multirow}
\usepackage{graphicx}
\usepackage[normalem]{ulem}
\usepackage{caption} 
\usepackage{subcaption}
\usepackage{dsfont}
\begin{document}

\title{ST-MTM: Masked Time Series Modeling with Seasonal-Trend Decomposition for Time Series Forecasting}


\author{Hyunwoo Seo}
\email{ta57xr@unist.ac.kr}
\affiliation{%
  \institution{Ulsan National Institute of Science and Technology}
  \city{Ulsan}
  \country{Republic of Korea}
}

\author{Chiehyeon Lim}
\authornote{Corresponding author}
\email{chlim@unist.ac.kr}
\affiliation{%
  \institution{Ulsan National Institute of Science and Technology}
  \city{Ulsan}
  \country{Republic of Korea}
}

\renewcommand{\shortauthors}{Seo and Lim.}

\begin{abstract}
Forecasting complex time series is an important yet challenging problem that involves various industrial applications. Recently, masked time-series modeling has been proposed to effectively model temporal dependencies for forecasting by reconstructing masked segments from unmasked ones. However, since the semantic information in time series is involved in intricate temporal variations generated by multiple time series components, simply masking a raw time series ignores the inherent semantic structure, which may cause MTM to learn spurious temporal patterns present in the raw data. To capture distinct temporal semantics, we show that masked modeling techniques should address entangled patterns through a decomposition approach. Specifically, we propose ST-MTM, a masked time-series modeling framework with seasonal-trend decomposition, which includes a novel masking method for the seasonal-trend components that incorporates different temporal variations from each component. ST-MTM uses a period masking strategy for seasonal components to produce multiple masked seasonal series based on inherent multi-periodicity and a sub-series masking strategy for trend components to mask temporal regions that share similar variations. The proposed masking method presents an effective pre-training task for learning intricate temporal variations and dependencies. Additionally, ST-MTM introduces a contrastive learning task to support masked modeling by enhancing contextual consistency among multiple masked seasonal representations. Experimental results show that our proposed ST-MTM achieves consistently superior forecasting performance compared to existing masked modeling, contrastive learning, and supervised forecasting methods.

\end{abstract}

\ccsdesc[500]{Mathematics of computing~Time series analysis}
\ccsdesc[500]{Computing methodologies~Dimensionality reduction and manifold learning}

\keywords{Masked modeling, Time series forecasting, Seasonal-trend decomposition, Self-supervised learning}

\maketitle

\section{Introduction}
Time series forecasting has been widely applied to various industrial domains, such as energy consumption, traffic, and weather. However, it remains a challenging task due to the complex temporal patterns in time series (e.g., continuity, seasonality, and trend) \cite{cai2020traffic, hewage2020temporal}. Beyond the recent rise in supervised deep forecasting models \cite{informer, autoformer}, self-supervised learning has been actively explored to pre-train models to identify useful time series representations through pretext tasks on vast amounts of unlabeled data \cite{ts2vec, tstcc}. Meanwhile, masked modeling has become a promising pre-training paradigm in various fields, such as masked image modeling (MIM) in computer vision and masked language modeling in natural language processing (MLM) \cite{bert}. Accordingly, masked time-series modeling (MTM) has been proposed to extend masked modeling to time series analysis \cite{tst, timae}.

The objective of MTM is to model temporal dependencies through the reconstruction of masked segments based on the unmasked parts
\cite{tst, timae, simmtm, patchtst, tarnet}. However, real-world time series exhibits intricate temporal variations, where heterogeneous structured patterns are entangled \cite{timesnet}. As these salient temporal dependencies can be obscured deeply in mixed temporal patterns, simply masking portions of raw time series ignores the inherent semantic information of structured patterns and can cause MTM to learn spurious temporal dependencies manifest in the raw data (see Figure \ref{figure1}). To capture distinct temporal dependencies within time series, we propose that masked modeling technique for time series should address entangled patterns through a decomposition approach.

One intuitive way to disentangle complex temporal variations is through the utilization of seasonal-trend decomposition that defines a time series as the sum of seasonal and trend components with noise, which has been recently validated as effective in deep time series forecasting \cite{stl1, autoformer, dlinear}. Decomposition can guide the model to extract salient temporal patterns: according to the analysis of MTMs on the ETTh1 dataset, as shown in Figure \ref{figure1}, the MTM Transformer encoder on the raw time series produces an indistinguishable attention score distribution, whereas the attention maps on its trend and seasonal components reveal apparent temporal patterns. Specifically, each component exhibits a different temporal dependency. The semantic information of each time point in the trend component is mainly involved in its adjacent time points. The seasonal component, on another hand, often represents similar temporal variations at positions of its multiple inherent periods \cite{timesnet, autoformer}. Recent studies also suggest that masking semantically meaningful parts can guide the masked model to learn high-level representations \cite{semmae}. These findings imply that semantics-aware masking of each decomposed component may be effective to understand the intricate temporal relationships in masked time-series modeling. 

Based on this motivation, we propose ST-MTM, a novel Masked Time-series Modeling framework with Seasonal-Trend decomposition for time series forecasting. To effectively model complex temporal patterns in raw time series, ST-MTM incorporates a decomposition architecture in both masking and representation learning methods. ST-MTM involves seasonal-trend masking and representation learning of each component. For seasonal-trend masking, we introduce two methods: period masking for seasonal time series and sub-series masking for trend time series, which reflect the inherent temporal semantics of each component. These methods allow ST-MTM to learn seasonal and trend representations independently and integrate them through the proposed component-wise gating layer. Then, ST-MTM reconstructs the original time series from the masked seasonal and trend series. Additionally, we present contrastive learning to capture consistent contextual information on multiple masked seasonal series, assuming that different masked series contain similar global contexts \cite{ts2vec}. Empowered by this design, ST-MTM achieves state-of-the-art and comparable performance on nine time series forecasting benchmarks. The main contributions of our work are summarized as follows:

\begin{itemize}
\item Building upon existing MTM and deep time series forecasting methods, we identify the necessity of a decomposition approach for MTM to explicitly capture distinct temporal variations in time series components.
\item Specifically, we propose ST-MTM, a decomposition architecture for MTM. ST-MTM involves a seasonal-trend masking method that removes regions sharing similar semantic information in each component, posing an effective self-supervisory task to understand the different semantic relationships within each component. Furthermore, ST-MTM captures consistent global contexts of masked series through contextual contrastive learning. 
\item We evaluate ST-MTM on numerous time series benchmark datasets for forecasting, comparing it with state-of-the-art masked modeling methods, contrastive learning, and supervised forecasting methods with a decomposition architecture. We further validate the effectiveness of our seasonal-trend masking and representation learning through ablation studies. 
\end{itemize}

\section{Related work}
\subsection{Self-supervised Learning for Time Series}

Self-supervised learning has emerged as an important research area with its capacity to learn meaningful representations from unlabeled data across various domains \cite{ecgssl, data2vec, simclr}. Through pre-training with pretext tasks \cite{deepclustering}, self-supervised learning has successfully enabled the capture of underlying structures within data and identified effective representations for downstream tasks. Recently, contrastive learning has gained attention as an effective pretext task \cite{simclr}, aiming to learn a representation space where positive pairs are pulled closer and negative pairs are pushed apart. TS2Vec \cite{ts2vec} uses hierarchical contrastive methods to learn the granularity of temporal contexts. CoST \cite{cost} proposes contrastive learning in both the time and frequency domain for learning seasonality-trend representations. LaST \cite{last} achieves the disentanglement of seasonal-trend representations using variational inference. While contrastive learning has shown good performance in high-level tasks \cite{tstcc, ts2vec}, such as time series classification, instance-wise contrasting inherently has difficulties in learning intricate temporal dependencies within time series, which are crucial for time series forecasting \cite{simmtm}.


\begin{figure}[t]
    \centering
    \begin{subfigure}{\columnwidth}
        \centering
        \includegraphics[width=0.49\columnwidth]{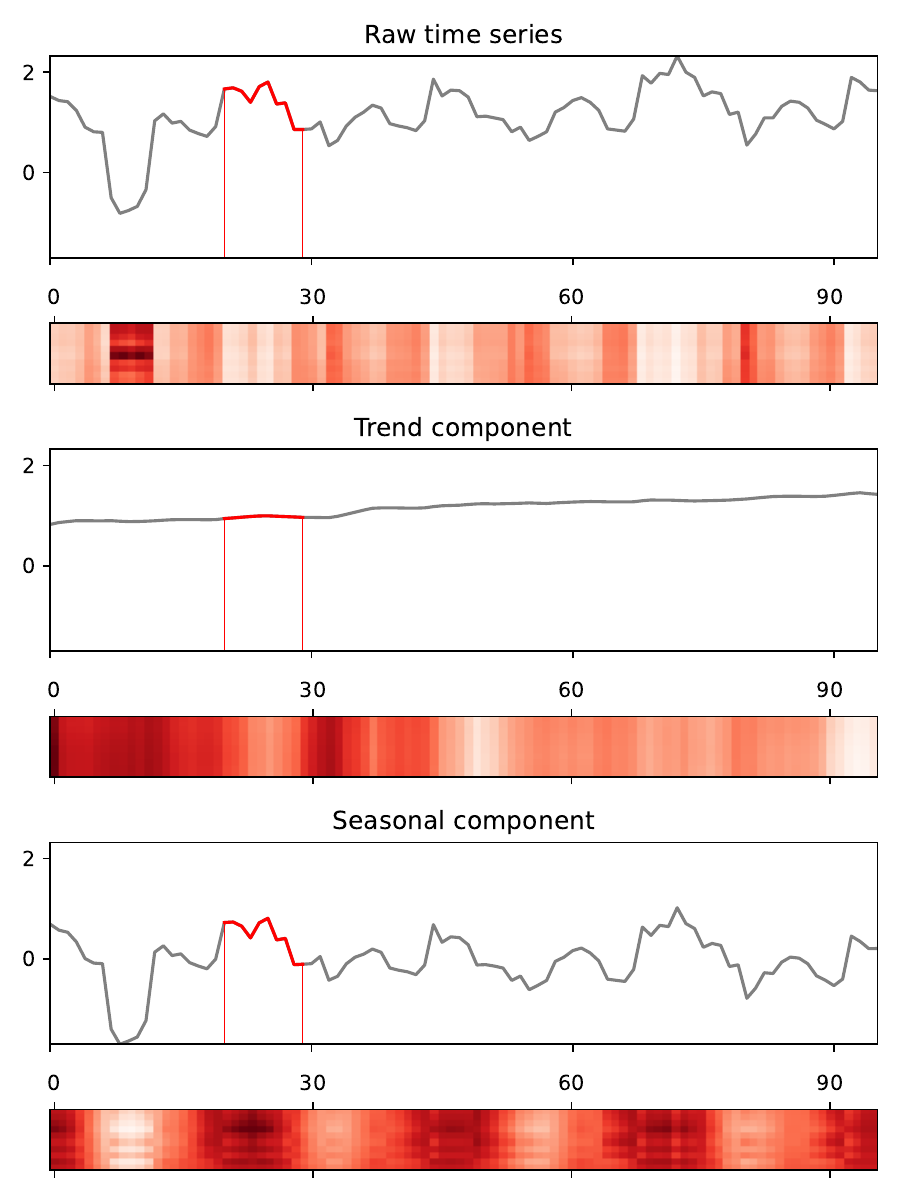}
        \includegraphics[width=0.49\columnwidth]{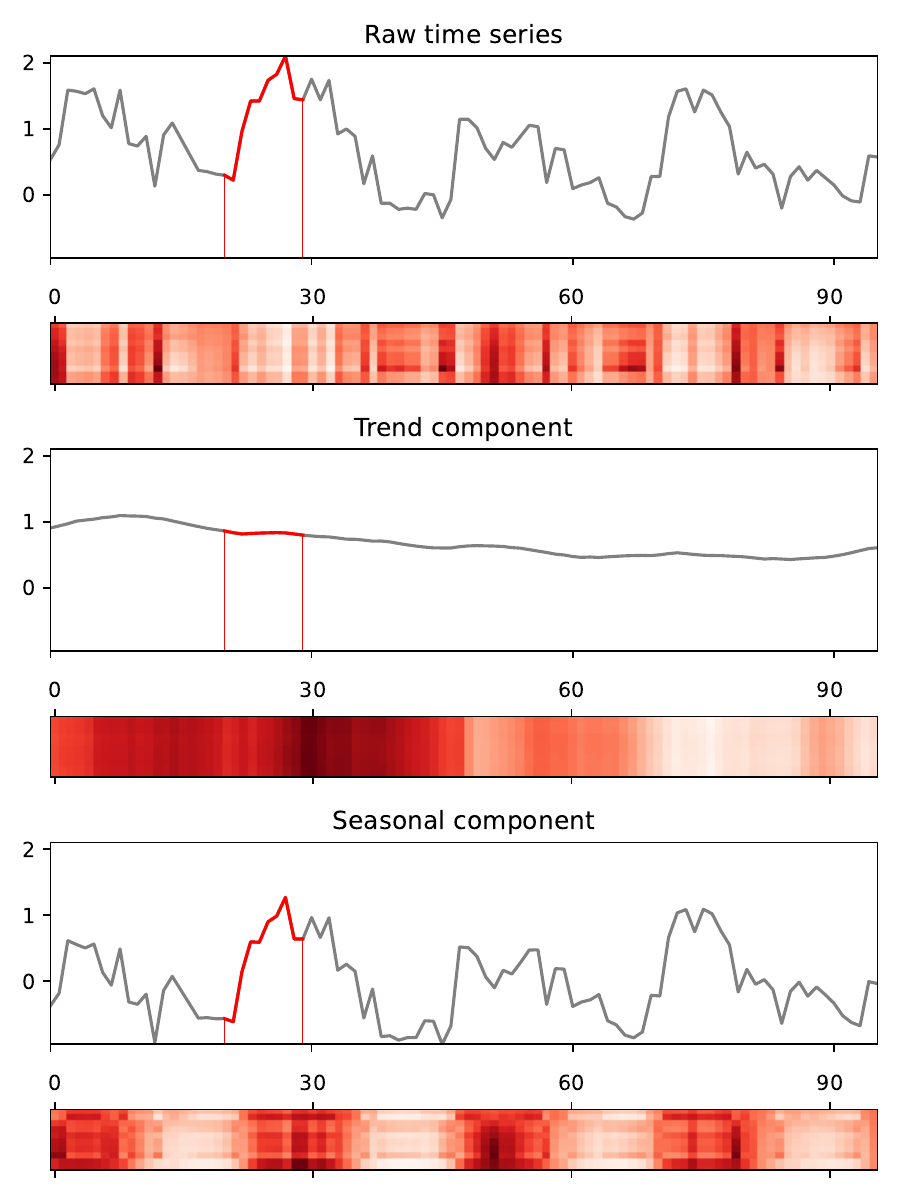}        
    \end{subfigure}
    \vspace{-5mm}
    \caption{The attention score distributions of the MTM Transformer encoder (PatchTST with patch length 1) pre-trained on the ETTh1 dataset through the reconstruction of the raw time series are depicted. A darker color indicates a higher attention score. We can observe that MTM learns spurious temporal patterns from the raw time series, whereas the attention map of its trend and seasonal components exhibit clear and distinct temporal patterns. This demonstrates that seasonal-trend components have different temporal dependencies.}
    \label{figure1} 
\end{figure}

\begin{figure*}[t]
  \centering
  \includegraphics[width=\textwidth]{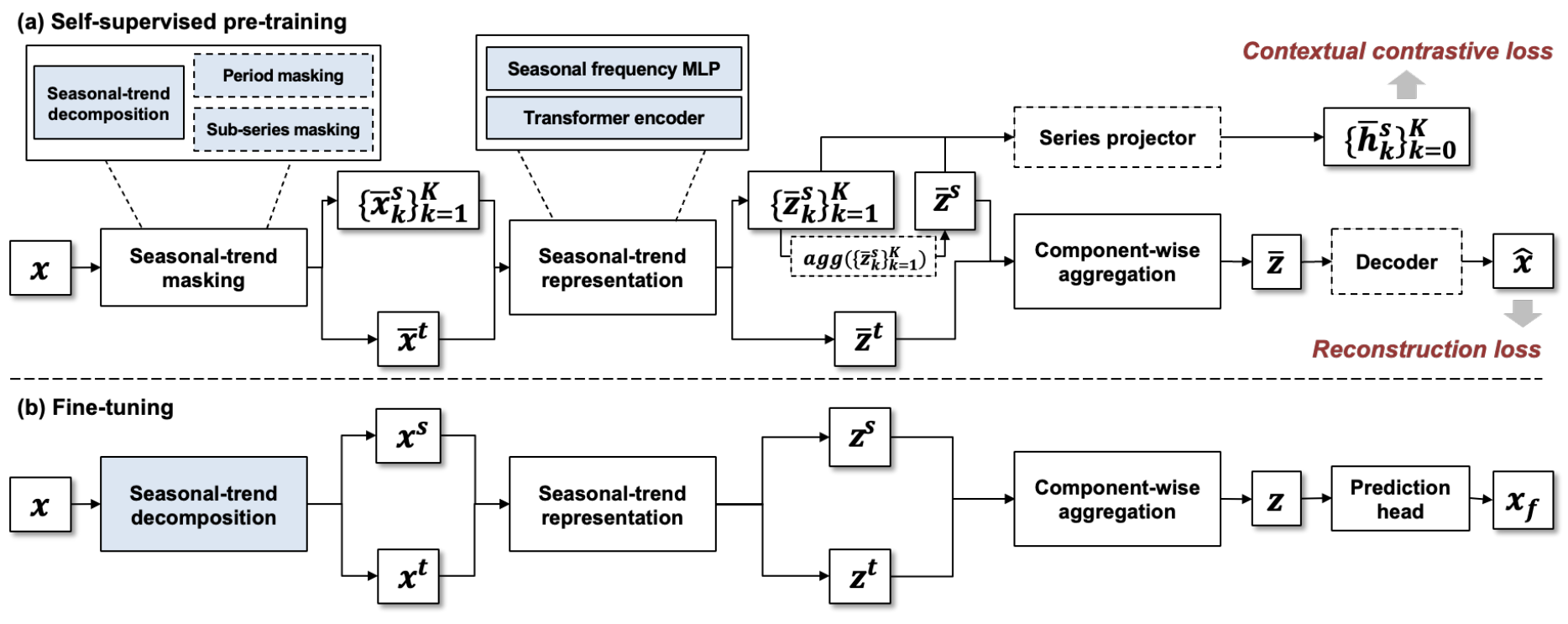}
  \caption{Overall architecture of ST-MTM. The ST-MTM architecture includes self-supervised pre-training and fine-tuning stages. Dashed boxes indicate modules used only during the self-supervised pre-training stage.}
  \vspace*{-3.5mm}
  \label{figure2}
\end{figure*}

\subsection{Masked Time-Series Modeling}
Masked time-series modeling has been actively explored as a self-supervised method for temporal dependency modeling \cite{tst}. In the general structure of MTM, the masking design is a key phase that determines the properties of representation. TST and Ti-MAE \cite{tst, timae} randomly mask a portion of time points in the raw data and PatchTST \cite{patchtst} applies masking to patches (i.e., sub-series) of the raw data to encode local semantic information. TARNet \cite{tarnet} designs a task-aware masking by using the self-attention score distribution from the end-task to improve end-task performance. SimMTM \cite{simmtm} generates multiple masked time series to effectively model the data manifold in the representation space. TimeSiam \cite{timesiam} reconstructs randomly masked series by extracting relevant temporal information from sub-series at previous time steps. However, masking the raw time series can lead MTMs to learn spurious dependencies present in the raw data. As multiple variations are intricately overlapped in the raw time series, masking the raw time series cannot consider distinct properties involved in various temporal patterns, such as trend and seasonality (refer to Figure \ref{figure1}). As such, we suggest that MTM should disentangle mixed temporal patterns within time series to capture the distinct dependencies that each temporal pattern presents.



\subsection{Seasonal-Trend Decomposition for Time Series Forecasting}
The time series decomposition breaks down a complex time series into several components, each representing distinct temporal patterns \cite{stl1}. Recent works have leveraged the decomposition strategy in deep learning approaches to effectively unravel intricate temporal patterns within time series and achieve interpretability. Autoformer \cite{autoformer} proposes decomposition blocks as inner operators in Transformers to empower the deep forecasting models through progressive decomposition. FEDformer and ETSformer \cite{fedformer, etsformer} utilize frequency-domain operations to enhance seasonal-trend decomposition. DLinear \cite{dlinear} extracts the trend and seasonal parts from raw data and applies a one-layer linear layer to each part to predict the future horizon. Further, SCNN \cite{scnn} decomposes time series into more granular components to model the detailed interactions among these components. Meanwhile, despite these studies demonstrating the significance of the decomposition architecture, there has been no attempt to incorporate this architecture into masked time-series modeling for capturing complex temporal variations. Our proposed ST-MTM integrates the decomposition scheme with time series masking and representation learning to extract salient temporal dependencies obscured in the raw time series.



\section{ST-MTM}
The pre-training process of ST-MTM and its essential modules are depicted in Figure \ref{figure2}. As shown, the pre-training of ST-MTM involves seasonal-trend masking, seasonal-trend representation learning, reconstruction, and contextual contrastive learning. The code is available at the official repository\footnote{\url{https://github.com/hwseo95/st-mtm}}.

\subsection{Seasonal-Trend Masking}
We propose a decomposed masking strategy for the seasonal and trend components of each time series. Specifically, given $\{x_{i}\}_{i=1}^N$ as a mini-batch, a time series $x_i \in \mathbb{R} ^{L\times C}$ comprises $L$ timestamps and $C$ variables. For each $x_i$, we generate a masked trend time series $\bar{x}_i^t$ and a set of $K$ masked seasonal time series $\{\bar{x}_{i,k}^s\}_{k=1}^K$. Hereafter, we omit the superscript  $\bar{}$ and the subscript ${}_i$ for simplification. Initially, we use the mean-normalized time series as input by subtracting the average value of all time series in a batch to remove the offset from data \cite{norm}, and adding it back to the final output of ST-MTM. Then, we extract trend time series $x^t$ and seasonal time series $x^s$ from $x$ by adopting the moving average operation as:
\begin{equation}
\begin{gathered}
    x^t = avgpool(padding(x))  \\   
    x^s = x - x^t
\end{gathered}
\label{movingaverage}
\end{equation}
where $x^s$, $x^t \in \mathbb{R}^{L \times C}$ denote the extracted seasonal and trend time series, respectively. We apply the padding operation to maintain the length unchanged after the moving average as in \cite{autoformer}.

\subsubsection{Period masking for seasonal time series}
As demonstrated in  Figure \ref{figure1}, it has been experimentally shown that seasonal components exhibit similar periodic behavior at multiple lag positions. Based on this observation, we propose a period masking strategy that considers inherent multi-periodicity. Initially, we calculate the autocorrelation of $x^s$. Drawing from the theory of stochastic processes \cite{stochastic1, stochastic2}, we derive the autocorrelation for a real discrete-time process $\{x_t\}$ using the following equation:
\begin{equation}
  r(\tau)=\mathbb{E}(x_{t}x_{t-\tau})
\end{equation}
$r(\tau)$ represents the similarity between time lag positions at $\tau$. To discover periods, we choose the most probable $K$ period length $\tau_1, ..., \tau_K$ as the time lags with the top-$K$ autocorrelations:
\begin{equation}
 \{\tau_1, ...,\tau_K\} = argTopk_{\tau \in \{1,..,L\}} \left( Avg(r_{xx}(\tau)) \right)
\end{equation}
where $K$ is the hyper-parameter. The periods identified through autocorrelation enable us to discover segments affected by the variations of adjacent periods. For efficient autocorrelation computation, we calculate $r_{xx}(\tau)$ by using Fast Fourier Transform (FFT) based on the Wiener-Khinchin theorem \cite{theorem, autoformer}.

For each $\tau_i$, we randomly sample a sub-series of length $l$ in $x^s$ as an anchor. Then, we mask all sub-series at positions that are n multiples of the period away from the anchor sub-series. Finally, we have a set of $K$ masked seasonal series $\{\bar{x}_k^s \}_{k=1}^K$ based on the inherent periods. 

\begin{figure}[t]
  \centering
  \includegraphics[width=\linewidth]{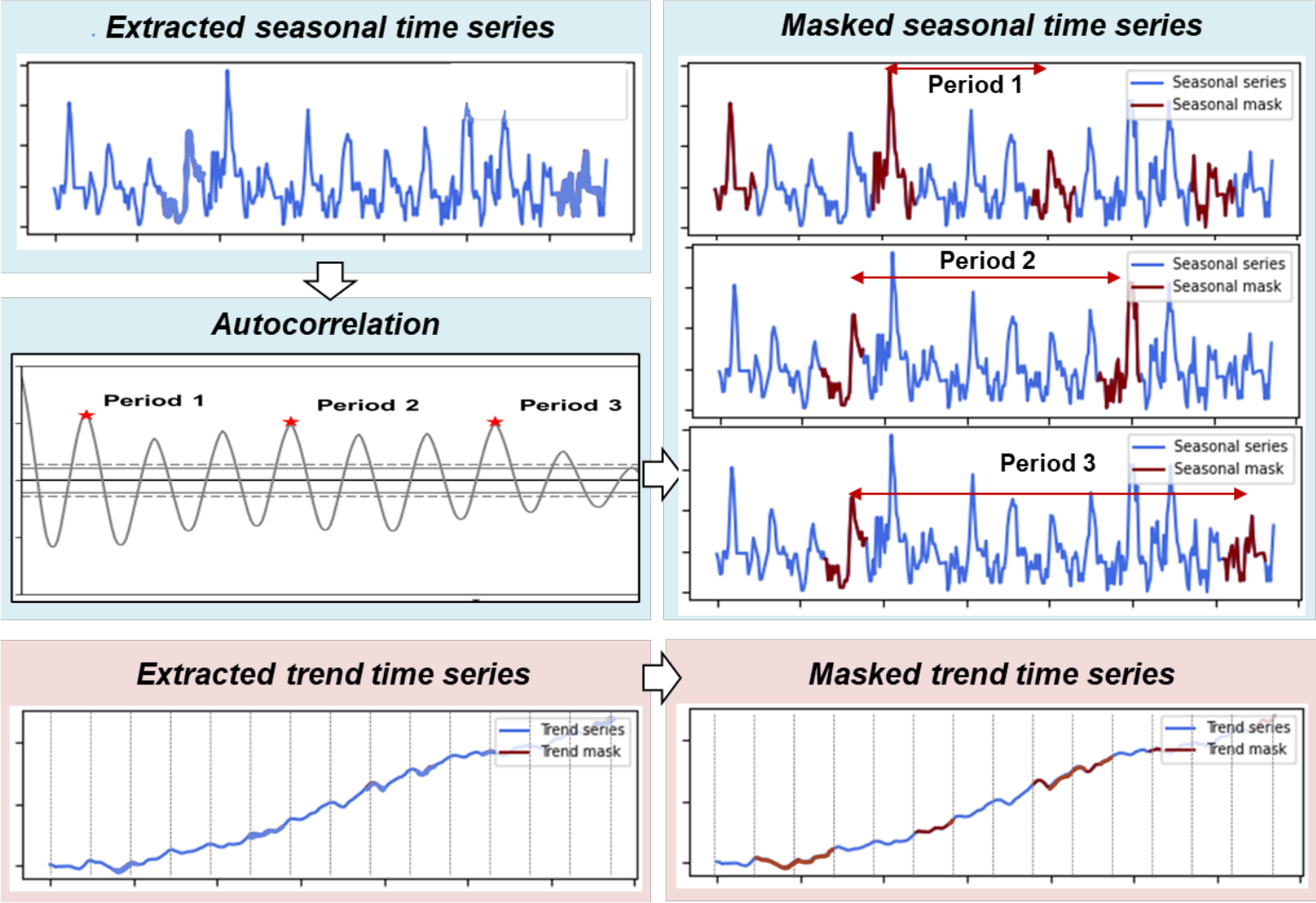}
  \caption{Seasonal-trend masking consists of period masking on a seasonal time series that masks related sub-series based on estimated periods with the K largest autocorrelation, and random sub-series masking on trend time series.}
  \vspace{-2mm}
  \label{figure3}
\end{figure}

\subsubsection{Sub-series masking for trend time series}
The semantic information of each time point in a trend primarily relates to its adjacent time points. However, masking at the level of single time steps can be easily inferred by interpolating with the preceding or succeeding time values without high-level understanding of the local semantic information \cite{patchtst}. Therefore, we introduce sub-series masking for trend time series to mask sub-series with similar temporal patterns. Inspired by \cite{patchtst}, each channel in $x^t$ is divided into non-overlapping sub-series of length $l$. Here, even if the length of the last segment is not equal to $l$, we still retain the last segment, resulting in a total number of sub-series $n_s=\left \lceil \frac{L}{l} \right \rceil$. We then randomly mask $p$ of $n_s$ segments for each channel. This decomposed seasonal-trend masking strategy explicitly separates the different temporal patterns in the masking paradigm, posing a challenging self-supervised task as it removes regions sharing similar semantic information and temporal dependencies.

\subsection{Seasonal-Trend Representation}
To obtain a time series representation from masked seasonal and trend time series, ST-MTM encodes the representation of each component independently, and aggregates them through a learnable aggregation layer. 

\begin{figure}[t]
  \centering
  \includegraphics[width=\linewidth]{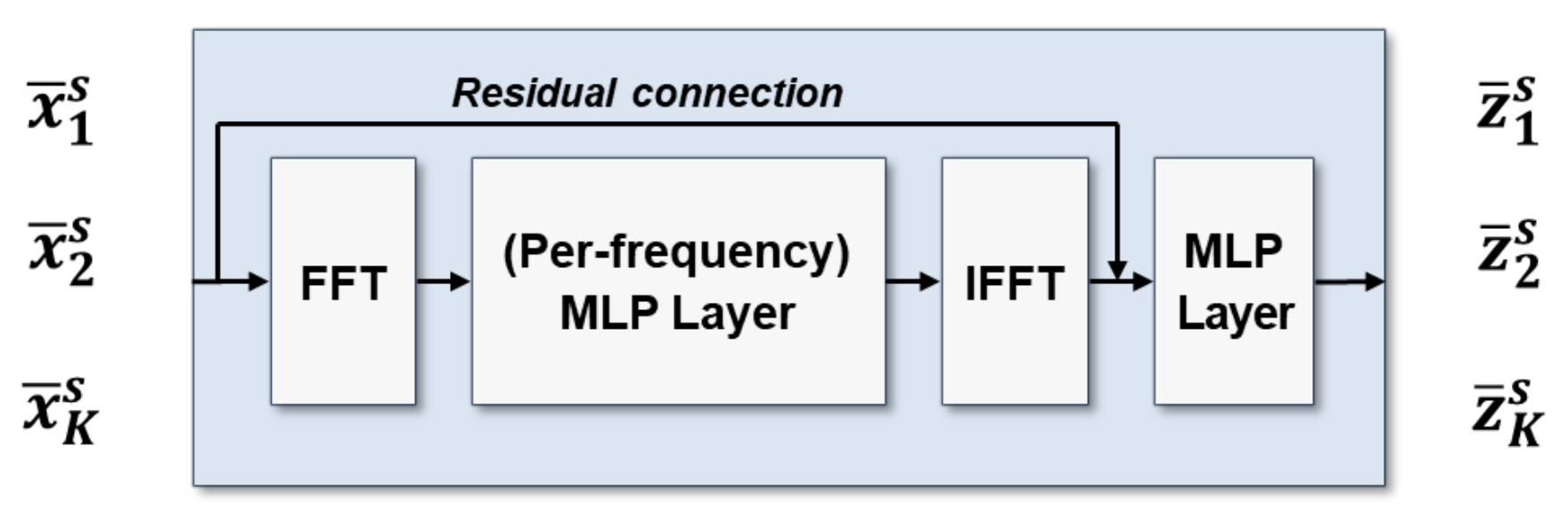}
  \vspace{-3mm}
  \caption{Seasonal frequency MLP}
  \label{figure4}
\end{figure}

\subsubsection{Encoding seasonal series}
To obtain a seasonal time series representation from a set of multiple masked seasonal series, ST-MTM first encodes each masked seasonal series into point-wise representations. The seasonal component of a time series exhibits multiple periodic properties, generated from its constituent frequencies \cite{timesnet}. To effectively capture the periodic information in masked seasonal time series, we propose the Seasonal Frequency MLP (SFM) as the encoder for seasonal time series. 

SFM consists primarily of FFT to convert time-domain input series into the frequency domain, a per-frequency MLP, and an inverse FFT, which maps the frequency-domain inputs back to the time domain. We initially embed the raw inputs into deep features by a learnable embedding layer $x^s_{emb}=Emb(x^s) \in \mathbb{R}^{L \times d_{model}}$. Then, the FFT transforms the time-domain representation $x^s_{emb}$ into frequency domain, $\mathcal{F}(x^s_{emb})\in \mathbb{C}^{F\times d_{model}}$, where $F=\left \lfloor L/2 \right \rfloor + 1$ is the number of frequencies. Utilizing FFT facilitates the decomposition of a time series into its constituent frequencies  \cite{frets}, aiding in the identification of seasonal patterns within the data. Subsequently, the per-frequency MLP layer performs an affine transformation and applies an activation function for each frequency. An inverse FFT then reverts the frequency domain representations back to the time domain as follows:
\begin{equation}
 z_{p, q}^s = \mathcal{F}^{-1}\left ( \sigma ( \sum_{p=1}^d W^f_{p,j,q} \mathcal{F}(x^s_{emb})_{p,j} + B^f_{p,j})   \right )
\end{equation}
where $W^f\in\mathbb{C}^{F \times d_{model} \times d_h}$ and $B^f\in\mathbb{C}^{d_{model} \times d_h}$ are the complex-valued parameters in per-frequency MLP layers. Following this, residual connection and MLP layers are applied as:
\begin{equation}
 z^s = \sigma (W(z^s + W^{rc}x^s_{emb})+B)
\end{equation}
where $W\in\mathbb{R}^{d_h \times d_{model}}$ and $B\in\mathbb{R}^{d_h \times d_{model}}$ are the parameters in the MLP layers in the time domain, and $W_{rc}\in\mathbb{R}^{d_{model} \times d_h}$ are the parameters for transformation in the residual connection. The ReLU activation is used in MLP layers in both the time and frequency domains. Consequently, the final output of SFM for the masked seasonal series $\{\bar{x}_k^s\}_{k=1}^K$ is the point-wise representations of the masked seasonal series $\{\bar{z}_k^s\}_{k=1}^K$.

\subsubsection{Adaptive aggregation of masked seasonal series}
Inspired by the approach in \cite{autoformer}, we aggregate $K$ masked seasonal representation using the autocorrelation $r$. $r$ can indicate the strength of the estimated periods on the time series, reflecting the significance of each masked seasonal series. We aggregate the masked seasonal representations based on the following calculation:
\begin{equation}
\begin{gathered}
 \hat{r}(\tau_1),..., \hat{r}(\tau_K) = Softmax ({r}(\tau_1), ..., {r}(\tau_K)) \\
 \bar{z}^s = \sum_{k=1}^K \hat{r}(\tau_k) \cdot  \bar{z}_k^s
\end{gathered}
\end{equation}
For conciseness, we denote $\bar{z}^s=\bar{z}^s_0$. As each masked seasonal series represents the distinctive temporal pattern of each period, the aggregated masked seasonal series adaptively reflects the semantic information of multiple periodic variations.

\subsubsection{Learning contextual representation of masked seasonal series} \label{contextual representation}
The instance-wise representation of the masked seasonal series $\{\bar{z}_k^s\}_{k=0}^K$ is learned through a series projector, which can be formulated as:
\begin{equation}
\{ \bar{h}_k^s \}_{k=0}^K = \mathbf{Projector_S}(\{\bar{z}_k^s\}_{k=0}^K)
\end{equation}
where $\bar{h}_k^s \in \mathbb{R}^{1 \times d_{model}}$. We employ a simple linear layer along the temporal dimension as the series projector to obtain instance-wise representations that capture the contextual information of the series. The output representations are used for contextual contrastive learning during pre-training, which will be depicted in Section \ref{contrastive loss}

\subsubsection{Trend series representation}
Following the trend encoder, we obtain point-wise representations of the trend time series $\bar{z}^t \in \mathbb{R}^{L \times d_{model}} $. For the trend encoder, we utilize Transformer, a standard architecture for learning representations of time series in masked modeling \cite{tst, patchtst}. Transformer is capable of simultaneously considering the long contexts of an input sequence and learning to represent each time point through a multi-head attention mechanism. While the trend component of a time series encapsulates the long-term progression, the Transformer is adept at modeling these long-term temporal variations.

\subsubsection{Component-wise aggregation}
By employing a decomposition scheme for masking strategies and encoder architectures, our goal is to integrate the decomposed components in the representation space, thereby capturing the temporal patterns of both seasonal and trend parts simultaneously. Specifically, we aim to obtain $z \in \mathbb{R}^{L \times d_{model}}$, which aggregates $z^t$ and $z^s$. While simple aggregation functions like addition and concatenation might not reflect the interaction of representations, we use a component-wise gating layer. The gating layer guides the model to learn the relative influence of seasonal and trend components at each timestamp:
\begin{equation}
\begin{gathered}
z_t = a_t \cdot z^t_t + b_t \cdot z^s_t \\
[a_t, b_t] = Softmax\left( g([z^t_t, z^s_t]) \right)
\end{gathered}
\end{equation}
where $t \in \{1,...,L\}$. A linear layer is utilized for $g(\cdot)$, facilitating a dynamic weighting that adaptively balances the influences of the seasonal and trend components to the final representation.


\subsection{Objective Function}
\subsubsection{Reconstruction loss}
As part of a self-supervised pre-training task, ST-MTM performs a reconstruction task, which is the standard pre-training paradigm in masked modeling. The reconstruction loss is formulated as:
\begin{equation}
L_{rec} = \sum_{i=1}^N ||x_i - \hat{x}_i||^2_2
\end{equation}
In this context, a reconstruction of the original time series $x_i$ is achieved with $\hat{x}_i=\mathbf{Decoder}(\bar{z}_i)$. We utilize a simple linear layer on the channel dimension for $\mathbf{Decoder}(\cdot)$. 

\subsubsection{Contextual contrastive loss} \label{contrastive loss}
The period masking on the seasonal component generates multiple masked seasonal series, all of which are considered the augmentations of the seasonal component. Thus, we expect these masked seasonal series to possess identical contextual information regarding the seasonal component. In addition, frequency-domain MLP layers in SFM can be viewed as global convolutions within the time domain, facilitating the recognition of global temporal dependencies \cite{frets}. 

To enhance the contextual consistency among the multiple masked seasonal representations, we introduce a contextual contrastive loss. Given $\{x_{i}\}_{i=1}^N$ as a mini-batch and the instance-wise seasonal representations for each $x_i$, $\mathbf{H}_i = \{\bar{h}_{i, k}^s \}_{k=0}^K$, we designate the aggregated time series representation $\bar{h}_{i,0}^s$ as the anchor, and the $K$ other masked seasonal representations as positive pairs. The contextual contrastive loss is then defined as:
\begin{equation}
L_{cl} = -\frac{1}{NK}\sum_{i=1}^N \sum_{k=1}^K \left(log \frac{exp(\bar{h}_{i,0}^s \cdot \bar{h}_{i,k}^s / \tau)}{\sum_{j=1}^N\sum_{k=0}^K  \mathds{1}_{\left[i\neq j \right]}exp(\bar{h}_{i,0}^s \cdot \bar{h}_{j,k}^s/\tau) }\right)
\end{equation}
Instance-wise representations of seasonal components from other time series in the same batch are used as negative samples. Contextual contrastive learning enhances the robustness of learned representations against disrupted seasonal patterns by aligning multiple masked seasonal representations closely.

The overall loss of ST-MTM is the combination of the reconstruction and contextual contrastive losses as follows:
\begin{equation}
L = L_{rec} + \alpha L_{cl}
\end{equation}
where $\alpha$ is the hyper-parameter that controls the weight of the contextual contrastive loss.

\begin{table*}[t]
\caption{Multivariate forecasting results compared with self-supervised methods in in-domain forecasting scenarios. We fix the input length $L=336$ and all the results are averaged from 4 different prediction lengths, that is \{96, 192, 336, 720\}. For ILI, $L=36$ and results are averaged over \{12, 24, 36, 48\}. The best results are in bold and the second best results are underlined. Baselines with * are models adopting a decomposition architecture.}
\label{table 1}
\vspace{-2.5mm}
\resizebox{\textwidth}{!}{%
\begin{tabular}{cc|cccccccccccccccc}
\hline
\multicolumn{2}{c|}{Models}  & \multicolumn{2}{c}{ST-MTM} & \multicolumn{2}{c}{SimMTM} & \multicolumn{2}{c}{PatchTST} & \multicolumn{2}{c}{TARNet} & \multicolumn{2}{c}{Ti-MAE}& \multicolumn{2}{c}{TS2Vec} & \multicolumn{2}{c}{CoST *} & \multicolumn{2}{c}{LaST *} \\ \cline{3-18} 
\multicolumn{2}{c|}{Metrics} & MSE & MAE & MSE & MAE & MSE & MAE & MSE & MAE & MSE & MAE & MSE & MAE & MSE & MAE & MSE & MAE \\ \hline
\multicolumn{2}{c|}{ETTh1} & \textbf{0.413} & \textbf{0.429} & 0.435 & 0.444 & \underline{0.424} & \underline{0.432} & 1.089 & 0.822 & 1.030 & 0.791 & 0.901 & 0.709 & 0.740 & 0.639 & 0.567 & 0.524 \\ 
\multicolumn{2}{c|}{ETTh2} & \textbf{0.344} & \textbf{0.388} & \underline{0.359} & \underline{0.396} & 0.363 & 0.399 & 2.312 & 1.273 & 2.632 & 1.290 & 2.152 & 1.163 & 1.628 & 1.002 & 0.956 & 0.700 \\ 
\multicolumn{2}{c|}{ETTm1} & \underline{0.350} & \underline{0.383} & 0.356 & 0.387 & \textbf{0.343} & \textbf{0.379} & 0.805 & 0.688 & 0.547 & 0.540 & 0.706 & 0.601 & 0.489 & 0.492 & 0.388 & 0.402 \\ 
\multicolumn{2}{c|}{ETTm2} & \textbf{0.253} & \textbf{0.315} & 0.267 & 0.326 & \underline{0.262} & \underline{0.322} & 1.507 & 0.982 & 1.996 & 1.056 & 0.982 & 0.731 & 0.843 & 0.672 & 0.408 & 0.405 \\ 
\multicolumn{2}{c|}{Weather} & \textbf{0.230} & 0.276 &  \underline{0.232} & 0.269 & 0.234 & \underline{0.268} & 0.270 & 0.388 & 0.312 & 0.381 & 1.823 & 1.001 & 1.112 & 0.798 & 0.234 & \textbf{0.267} \\ 
\multicolumn{2}{c|}{Electricity} & \textbf{0.170} & \underline{0.273} & 0.174 & 0.274 & \textbf{0.170} & \textbf{0.264} & 0.366 & 0.433 & 0.331 & 0.429  & 0.359 & 0.424 & 0.200 & 0.300 & 0.186 & 0.274 \\ 
\multicolumn{2}{c|}{PEMS08} & \textbf{0.204} & \underline{0.305} & 0.289 & 0.365 & \underline{0.223} & \textbf{0.301} & 0.299 & 0.367 & 0.300 & 0.399 & 0.244 & 0.332 & 0.268 & 0.374 & 0.249 & 0.353 \\ 
\multicolumn{2}{c|}{ILI} & \underline{2.757} & \underline{1.062} & 3.120 & 1.192 & \textbf{2.264} & \textbf{0.925} & 6.255 & 1.746 & 3.595 & 1.313 & 3.347 & 1.175 & 2.841 & 1.113 & 3.283 & 1.141 \\ 
\multicolumn{2}{c|}{Solar} & \textbf{0.195} & 0.271 & 0.241 & 0.285 & \textbf{0.195} & \underline{0.243} & 0.231 & 0.300 & \underline{0.218} & 0.301 & 0.237 & 0.312 & 0.219 & 0.277 & 0.237 & \textbf{0.229}\\ \hline
\end{tabular}%
}
\end{table*}

\begin{table*}[t]
\caption{Multivariate forecasting results compared with decomposition-based supervised forecasting methods in in-domain forecasting scenarios. We fix the input length $L=336$ and all the results are averaged from 4 different prediction lengths, that is \{96, 192, 336, 720\}. For ILI, $L=36$ and results are averaged over \{12, 24, 36, 48\}. The best results are in bold and the second best results are underlined.}
\label{table 2}
\vspace{-2.5mm}
\begin{tabular}{cc|cccccccccccccc}
\hline
\multicolumn{2}{c|}{Models}  & \multicolumn{2}{c}{ST-MTM} & \multicolumn{2}{c}{SCNN} & \multicolumn{2}{c}{TimesNet} & \multicolumn{2}{c}{DLinear} & \multicolumn{2}{c}{Autoformer}& \multicolumn{2}{c}{FEDformer} & \multicolumn{2}{c}{ETSformer} \\ \cline{3-16} 
\multicolumn{2}{c|}{Metrics} & MSE & MAE & MSE & MAE & MSE & MAE & MSE & MAE & MSE & MAE & MSE & MAE & MSE & MAE \\ \hline
\multicolumn{2}{c|}{ETTh1} & \textbf{0.413} & \underline{0.429} & \underline{0.421} & \textbf{0.427} & 0.489 & 0.483 & 0.444 & 0.454 & 0.562 & 0.533 & 0.451 & 0.472 & 0.573 & 0.534   \\ 
\multicolumn{2}{c|}{ETTh2} & \textbf{0.344} & \textbf{0.388}  & \underline{0.348} & \underline{0.389} & 0.409 & 0.441 & 0.409 & 0.431 & 0.623 & 0.581 & 0.415 & 0.454 & 0.421 & 0.453\\ 
\multicolumn{2}{c|}{ETTm1} & \textbf{0.350} & \textbf{0.383} & 0.516 & 0.477 & 0.441 & \underline{0.430} & \underline{0.361} & \textbf{0.383} & 0.531 & 0.506 & 0.390 & \underline{0.430} & 0.692 & 0.566   \\ 
\multicolumn{2}{c|}{ETTm2} & \textbf{0.253} & \textbf{0.315}  & 0.286 & 0.341 & 0.294 & 0.341 & \underline{0.280} & \underline{0.338} & 0.393 & 0.424 & 0.330 & 0.381 & 0.316 & 0.372  \\ 
\multicolumn{2}{c|}{Weather} & \textbf{0.230} & \textbf{0.276}  & 0.249 & \underline{0.286} & 0.250 & 0.287 & \underline{0.245} & 0.299 & 0.395 & 0.433 & 0.325 & 0.371 & 0.292 & 0.353  \\ 
\multicolumn{2}{c|}{Electricity} & \underline{0.170} & 0.273 & 0.182 & \underline{0.271} & 0.200 & 0.301 & \textbf{0.169} & \textbf{0.267} & 0.243 & 0.346 & 0.228 & 0.342 & 0.211 & 0.326\\ 
\multicolumn{2}{c|}{PEMS08} & \textbf{0.204} & \underline{0.305} & 0.465 & 0.482 & \underline{0.212} & \textbf{0.264} & 0.348 & 0.423 & 0.311 & 0.375 & 1.077 & 0.860 & 0.343 & 0.421\\ 
\multicolumn{2}{c|}{ILI} & \underline{2.757} & \underline{1.062} & \textbf{2.556} & \textbf{0.976} & 4.088 & 1.391 & 2.873 & 1.189 & 3.618 & 1.348 & 3.368 & 1.290 & 2.990 & 1.148\\ 
\multicolumn{2}{c|}{Solar} & \textbf{0.195} & \underline{0.271} & \underline{0.216} & \textbf{0.270} & 0.228 & 0.274 & 0.253 & 0.314 & 0.781 & 0.640 & 0.245 & 0.338 & 0.719 & 0.668 \\ \hline 
\end{tabular}%
\end{table*}


\section{Experiments}
We extensively evaluate the proposed ST-MTM on nine benchmark datasets, covering various time series forecasting applications. We present the fine-tuning performance, which involves fine-tuning the prediction head and ST-MTM encoders in an end-to-end fashion. In addition to in-domain forecasting scenarios, we conduct experiments on cross-domain forecasting scenarios, where the model is pre-trained and fine-tuned on different datasets.

\subsection{Experimental Setup}
\subsubsection{Datasets}
The nine real-world benchmarks are summarized as follows. ETT consists of two hourly-level datasets (ETTh1, ETTh2) and two 15-minute-level datasets (ETTm1, ETTm2), which measure six power load features and oil temperature. Weather records 21 meteorological features every 10 minutes. Electricity contains  data on hourly electricity consumption for 321 customers. PEMS08 represents 5-minute traffic flows at 170 sensor locations. ILI contains weekly records of influenza-like illness patients. Solar collects the solar power production of 137 plants. We adopt the standard data pre-processing strategy in \cite{autoformer}, where the data in each variable is standardized. The statistics of the datasets are summarized in Appendix \ref{appendix1.1}.

\subsubsection{Baselines}
We compare ST-MTM with 13 baselines, comprising representative and state-of-the-art models in MTM, contrastive learning, and supervised forecasting methods with decomposition architecture. Baselines include TARNet \cite{tarnet}, Ti-MAE \cite{timae}, SimMTM \cite{simmtm}, and PatchTST \cite{patchtst} in MTM; TS2Vec \cite{ts2vec}, CoST \cite{cost}, and LaST \cite{last} in contrastive learning; and Autoformer \cite{autoformer}, FEDformer \cite{fedformer}, DLinear \cite{dlinear}, ETSformer \cite{etsformer}, TimesNet \cite{timesnet}, and SCNN \cite{scnn} in decomposition-based forecasting methods. CoST and LaST are contrastive learning methods that adopt a decomposition approach for seasonal-trend representation. PatchTST was originally proposed as for both supervised forecasting and self-supervised methods, but we chose self-supervised PatchTST for fair comparison to evaluate the effectiveness of ST-MTM in self-supervised learning.

\subsubsection{Implementation detail}
We adopt the channel independence design similar to SimMTM and PatchTST \cite{simmtm, patchtst}. The channel independence setting allows ST-MTM to focus on the temporal pattern in each univariate time series. We set the input length $L=336$ for all datasets except ILI, where $L=36$. We set the segment length for masking to 25, except for ILI, where it is set to 3 to maintain a similar number of segments in the input window. We set the masking ratio for the trend at 0.2, the number of masked seasonal series at 3, the temperature $\tau$ at 0.1, and the regularization parameter $\alpha$ at 0.5. We pre-train ST-MTM for 50 epochs and fine-tune it for 10 epochs, except for the Electricity and PEMS08 datasets, which are pre-trained for 10 epochs due to the time constraint. We implemented the baselines based on their official implementations and followed the configurations from their original papers. More implementation details are provided in Appendices \ref{appendix1.2} and \ref{appendix1.3}.

\subsection{Main Results}
We report the mean squared error (MSE) and mean absolute error (MAE) across a wide range of prediction lengths, \{96, 192, 336, 720\}, for all datasets except ILI, where \{12, 24, 36, 48\}. All experiments are repeated five times for each prediction length. We provide the complete results for all prediction lengths at our official repository.

\subsubsection{In-domain forecasting}
As shown in Table \ref{table 1}, ST-MTM outperforms the majority of self-supervised baselines, yielding competitive performance in some forecasting scenarios compared to PatchTST, which is the state-of-the-art MTM method. On average across all benchmarks, ST-MTM achieves the best score in 10 out of 18 forecasting scenarios and the second best score in six scenarios. Meanwhile, although TARNet presented remarkable performance in other downstream tasks such as classification and regression, its learnable masking with the attention score performs poorly in time series forecasting. ST-MTM also outperforms contrastive-based approaches. Although CoST and LaST adopt a decomposition approach in their frameworks, these methods report the poor forecasting performance. These results confirm the superior capability of ST-MTM in modeling complex temporal dependencies compared to other decomposition-based self-supervised methods.

Table \ref{table 2} demonstrates the superior performance of ST-MTM over decomposition-based supervised forecasting baselines across most datasets. ST-MTM achieves the best score in 11 out of 18 forecasting scenarios and the second best score in six scenarios. Specifically, ST-MTM outperforms the recent decomposition-based forecasting method, SCNN. ST-MTM also outperforms Transformer-based forecasting methods, Autoformer, FEDformer, and ETSformer, which involve the iterative decomposition of the time series through multiple decomposition blocks. We suggest that the simple seasonal-trend decomposition and semantics-aware masking in ST-MTM effectively capture heterogeneous temporal patterns of decomposed components. This approach appears more effective than granular component modeling in SCNN and progressive decomposition in decomposition-based Transformers.

\begin{table}[t]
\centering
\caption{Cross-domain forecasting results compared with self-supervised methods. The results are averaged from all prediction lengths \{96, 196, 336, 720\}.}
\vspace{-2mm}
\label{table transfer}
\resizebox{\columnwidth}{!}{
\begin{tabular}{cccccccc}
\hline
\multicolumn{2}{c}{Dataset} & \multicolumn{2}{c}{ST-MTM} & \multicolumn{2}{c}{SimMTM} & \multicolumn{2}{c}{PatchTST}
\\ \cline{3-8}
Source & Target & MSE & MAE & MSE & MAE & MSE & MAE \\ \hline
\multirow{2}{*}{ETTh1} & ETTh2 & \textbf{0.354} & \textbf{0.396} & 0.379 & 0.406 & 0.358 & 0.397 \\
& ETTm2 & \textbf{0.257} & 0.320 & 0.273 & 0.328 & 0.261 & \textbf{0.318} \\
\multirow{2}{*}{ETTm1} & ETTh2 & \textbf{0.350} & \textbf{0.398} & 0.395 & 0.416 & 0.360 & 0.397 \\
 & ETTm2 & \textbf{0.250} & \textbf{0.317} & 0.285 & 0.337 & 0.265 & 0.322 \\
 \multirow{2}{*}{ETTm2} & ETTh2 & \textbf{0.348} & \textbf{0.390} & 0.366 & 0.400 & 0.362 & 0.397 \\
 & ETTm1 & \textbf{0.350} & 0.383 & 0.448 & 0.425 & \textbf{0.350} & \textbf{0.382} \\ \hline
\end{tabular}
}
\end{table}

\subsubsection{Cross-domain forecasting}
We evaluate forecasting performance in cross-domain forecasting scenarios, where the model is pre-trained and transferred to different datasets. We select SimMTM and PatchTST as comparative baselines, as they demonstrate superior performance among self-supervised methods in in-domain forecasting scenarios. As shown in Table \ref{table transfer}, ST-MTM achieves superior forecasting performance in cross-domain scenarios, confirming the better transferability and robustness of the learned representations to mismatched frequencies between source and target datasets.

\subsection{Ablation Studies}
\subsubsection{Pre-training tasks}
\begin{figure}[t]
  \centering
  \includegraphics[width=\linewidth]{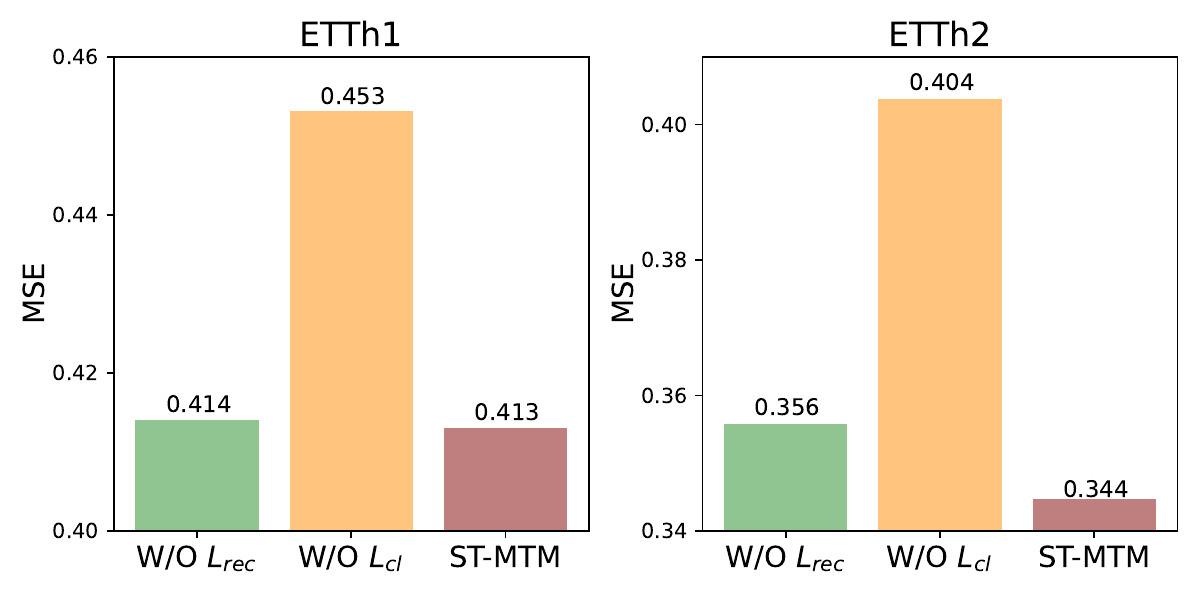}
  \vspace{-2mm}
  \caption{Ablation of ST-MTM on the reconstruction task ($L_{rec}$) and contextual contrastive learning task ($L_{cl}$) in time series forecasting. The results are averaged from 4 different prediction lengths, \{96, 192, 336, 720\}.}
  \label{figure5}
\end{figure}

We conduct an ablation study to demonstrate the effect of two pre-training tasks in ST-MTM, implemented through two parts of the training loss, $L_{rec}$ and $L_{cl}$. We removed each loss and recorded the final results (see Figure \ref{figure5}). The results show that both tasks are essential for forecasting. Here, $L_{cl}$ contributes more to the performance than $L_{rec}$. Contextual contrastive learning aligns masked seasonal representations of distinct temporal patterns from different periods within a seasonal component. As masking can be regarded as a data augmentation in contrastive learning \cite{ts2vec}, contextual contrastive learning guides the model to learn the robust semantic information within complex temporal variations. Therefore, we suggest that pre-training with contextual contrastive loss enhances forecasting performance on time series exhibiting periodic patterns.

\subsubsection{Seasonal-trend masking} \label{effect of masking}
Unlike conventional MTM methods, ST-MTM introduces seasonal-trend masking, which masks regions with similar semantic information. It aggregates multiple masked seasonal series based on autocorrelation and ensures consistent seasonal contexts in masked seasonal representations through contextual contrastive learning. To verify the effectiveness of our masking method and related modules, we conduct ablation studies on masking methods, the number of masked seasonal series and their aggregation, and consistency achieved through contextual contrastive loss. We selected two masking methods for comparison: one based on a random Bernoulli distribution and the other on a geometric distribution on both components. These methods randomly mask timestamps without considering the semantic information in the time series. For the number of masked seasonal series and their aggregation, we examined a single masked series without aggregation and multiple masked series with mean aggregation. Consequently, we define six scenarios for comparison with ST-MTM (see Table \ref{table3}). We set the masking ratio for random and geometric masking to 0.5, as proposed by SimMTM as the optimal masking ratio, and fixed the number of masked series at 3.

\begin{table}[h]
\centering
\caption{Different masking scenarios}
\label{table3}
\vspace{-2mm}
\resizebox{\columnwidth}{!}{%
\begin{tabular}{cccc}
\hline
\multirow{2}{*}{Name} & \multirow{2}{*}{Masking} & The number of masked series & Contextual \\
 &  & \& aggregation & consistency \\ \hline
R1 & Random & Single masked series without aggregation & X \\
R2 & Random & Multiple masked series with mean aggregation & X \\
R3 & Random & Multiple masked series with mean aggregation & O \\
G1 & Geometric & Single masked series without aggregation & X \\
G2 & Geometric & Multiple masked series with mean aggregation & X \\
G3 & Geometric & Multiple masked series with mean aggregation & O \\
ST & Period & Multiple masked series with adpative aggregation & O \\ \hline
\end{tabular}}
\end{table}

\begin{figure}[t]
  \centering
  \vspace{-2mm}
  \includegraphics[width=\linewidth]{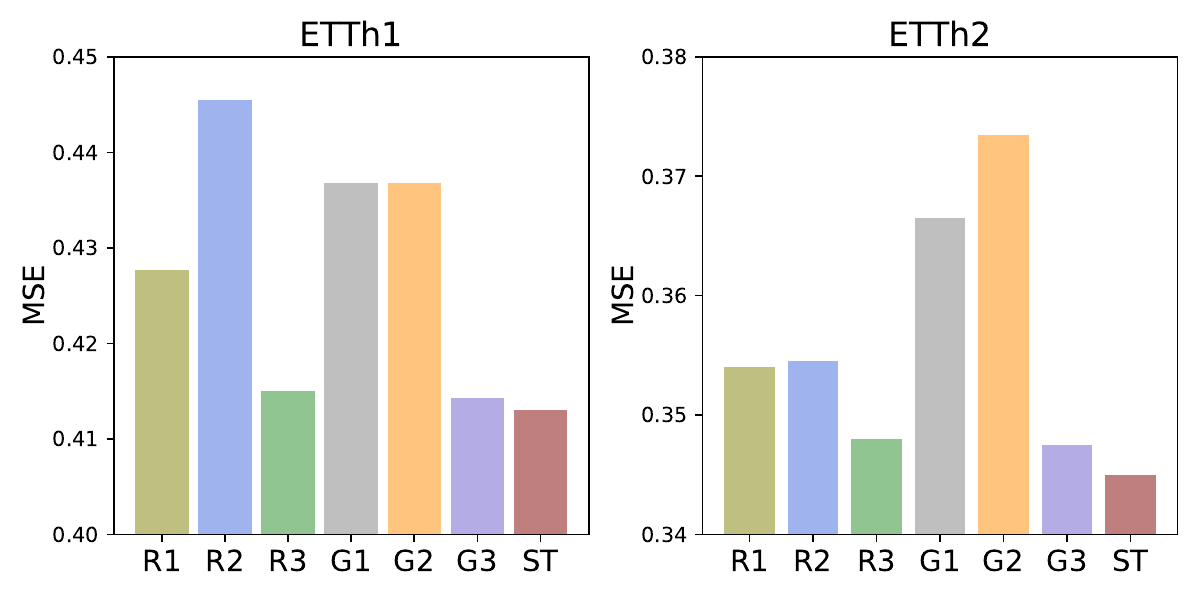}
  \caption{The MSE performance of seven masking scenarios. We report the average MSE score for all prediction length, \{96, 192, 336, 720\}.}
  \label{figure6}
\end{figure}

As shown in Figure \ref{figure6}, ST-MTM consistently outperforms other masking scenarios. It is observed that learning consistent seasonal contexts among masked seasonal series significantly improves the forecasting performance, regardless of the masking methods used (see R3, G3, and ST). Among these, sub-series masking and period masking of ST-MTM exhibit the best performance (see ST). This suggests that our design provides an effective pre-training task by removing regions sharing similar temporal information, which facilitates the understanding of complex temporal variations and enhances forecasting capabilities. Notably, the number of masked seasonal series does not have a positive impact on performance if these representations are not aligned (see MSE increase from R1 to R2, and from G1 to G2). Thus, we suggest that seasonal-trend masking, autocorrelation-based aggregation, and contextual contrastive learning are well-suited for capturing complex temporal patterns within the decomposition architecture of MTM. More ablation study results are available in Appendix \ref{appendix3}.

\subsection{Model Analysis}
\subsubsection{Component-wise gating layer}
\begin{figure}[t]
    \centering
    \begin{subfigure}{\columnwidth}
        \centering
        \includegraphics[width=0.49\columnwidth]{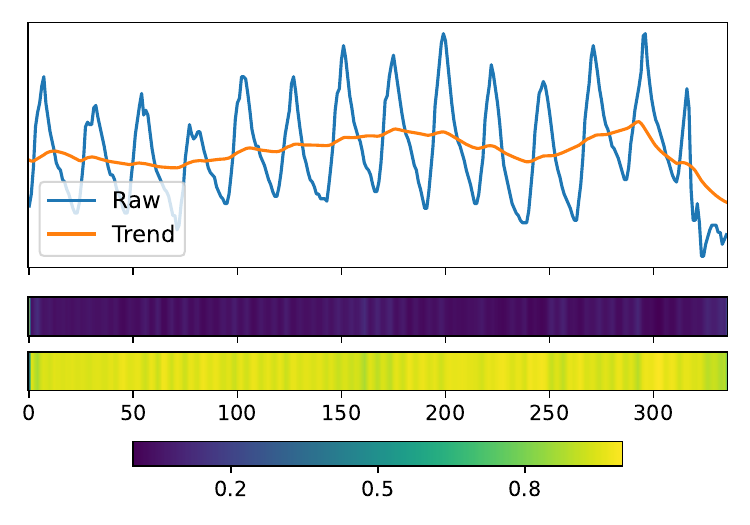}
        \includegraphics[width=0.49\columnwidth]{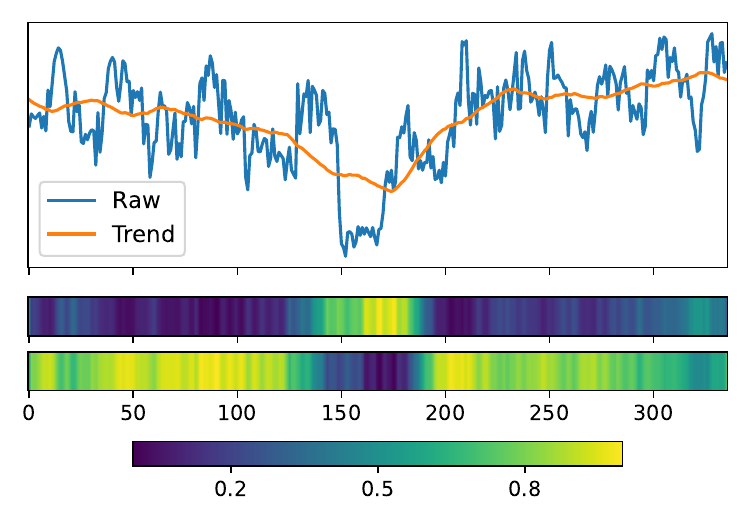}   
    \end{subfigure}
    \vspace{-5mm}
    \caption{Visualization of the outputs from the component-wise gating layer on the ETTh2 dataset. The orange line represents the trend component of the raw time series, extracted using a moving average with a kernel size of 50. The upper color bar indicates the weights assigned to the trend component, while the lower color bar indicates the weights assigned to the seasonal component.}
    \label{figure7} 
\end{figure}

Figure \ref{figure7} shows the outputs of the component-wise gating layer on the ETTh2 dataset, which determines the weights of seasonal and trend representations on the aggregated time series representation at each timestamp. When the time series exhibits the strong periodic patterns, the gating layer predominantly assigns high weights to the seasonal component. On the other hand, the gating layer assigns high weights to the trend component when the seasonal pattern is disrupted and long-term movement suddenly changes. These findings demonstrate that the learnable gating layer dynamically assigns the influence of each component on the entangled temporal patterns of the original time series. This adaptive gating mechanism enables ST-MTM to produce the robust representation to indistinct patterns, demonstrating beneficial for time series forecasting with scarce temporal patterns. For the detailed experiment, refer to Appendix \ref{appendix2.1}.

As shown in Figure \ref{figure7}, the aggregation of the gating mechanism adaptively extracts interactions among components, while standard aggregation methods such as averaging or concatenation reflect a fixed dependency between them and do not consider their relative influence. To validate the effect of the component-wise gating layer, we replace it with concatenation and averaging. Table \ref{table4} indicates that using the gating layer to aggregate seasonal and trend representations outperforms standard aggregation functions. ST-MTM decomposes time series as a pre-processing usage and encodes the separated components independently. However, ST-MTM facilitates information exchange through the gating layer, outperforming the existing decomposition-based methods such as DLinear and CoST that neglect interactions between the components.

\begin{table}[t]
\centering
\caption{Effect of the component-wise gating layer}
\vspace{-2mm}
\label{table4}
\begin{tabular}{ccccccc}
\hline
\multirow{2}{*}{Dataset} & \multicolumn{2}{c}{Gating layer} & \multicolumn{2}{c}{Concatenation} & \multicolumn{2}{c}{Average} \\ \cline{2-7} 
 & MSE & MAE & MSE & MAE & MSE & MAE \\ \hline
ETTh1 & \textbf{0.413} & \textbf{0.429} & 0.430 & 0.441 & 0.423 & 0.436 \\
ETTh2 & \textbf{0.344} & \textbf{0.388} & 0.371 & 0.404 & 0.347 & 0.393 \\ \hline
\end{tabular}
\end{table}

\begin{figure}[t]
  \centering
  \includegraphics[width=\linewidth]{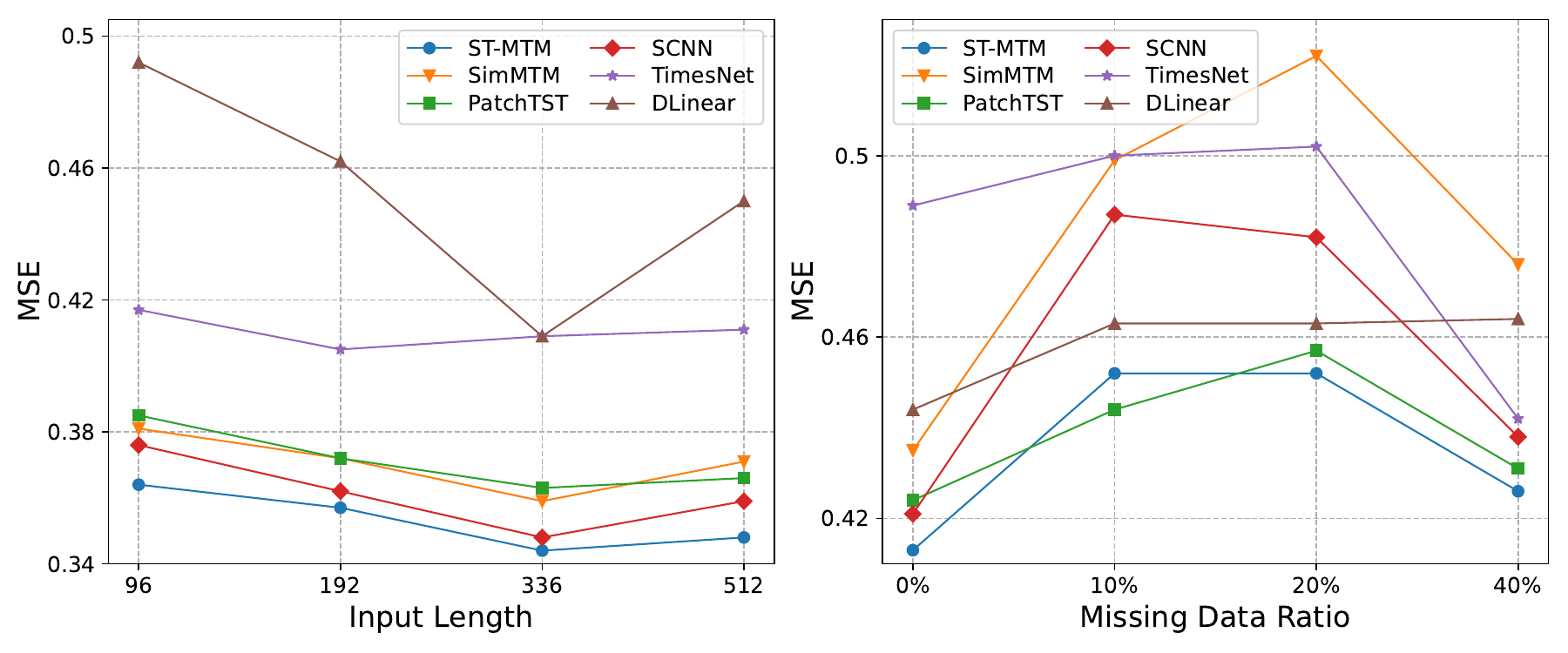}
  \vspace{-5mm}
  \caption{The left part shows MSE performance with varying input length on ETTh1. The right part shows MSE performance with various ratios of missing data on ETTh2. We report the average MSE score for all prediction lengths, \{96, 192, 336, 720\}. For this experiment, we use MTM and decomposition-based baselines that achieve the best MSE.}
  \label{figure8}
\end{figure}

\subsubsection{Various input length}
We study how ST-MTM can extract meaningful representations of seasonal and trend patterns for forecasting across various lengths of the look-back window. The left part of Figure \ref{figure8} shows the MSE performance for different input lengths \{96, 192, 336, 512\} on the ETTh1 dataset. We demonstrate that ST-MTM consistently outperforms other baselines at every input length. While decomposition-based forecasting baselines such as SCNN and DLinear exhibit a large MSE increase with shorter input lengths, our model maintains consistent performance with reduced input lengths compared to other baselines. These results confirm our model's capability to learn temporal dependencies of seasonal and trend components across various input lengths.

\subsubsection{Robustness analysis}
To evaluate model robustness, we construct a data corruption scenario of missing data. We randomly removed a portion of time points from both the train and test datasets, and then predict the original values in the test dataset. The less a model's performance degrades at the missing data, the more robust it is considered. As shown in the right part of Figure \ref{figure8}, ST-MTM consistently exhibits the smallest MSE performance among all models. We suggest that contextual contrastive learning enables the model to learn the robust representation against corrupted temporal patterns and extract the consistent contextual information from time series. These results demonstrate the superior robustness in the presence of missing values.

\section{Conclusion}
This study proposes ST-MTM, a masked time-series modeling framework with a seasonal-trend decomposition architecture designed to enhance temporal modeling capabilities. We identified that previous MTMs ignored the distinct temporal patterns generated by heterogeneous time series components, causing them to learn spurious temporal dependencies. The decomposition architecture of ST-MTM, applied in both masking and representation learning, enables our model to capture distinct temporal dependencies for seasonal and trend components within time series. Experimentally, ST-MTM demonstrates superior forecasting performance compared to recent self-supervised learning and decomposition-based forecasting methods. For future research, we aim to extend our work to masked time-series modeling using sequential decomposition, which could enhance the understanding of more detailed structured components in time series. Finally, while emerging time series foundation models have been trained using traditional self-supervised methods \cite{timer}, we believe that developing effective self-supervised methods for time series will further advance such models.

\bibliographystyle{ACM-Reference-Format}
\bibliography{reference}

\appendix

\section{Implementation details} \label{appendix1}
\subsection{Datasets} \label{appendix1.1}
The detailed descriptions of the benchmark datasets for time series forecasting are summarized in Table \ref{table 3}. The datasets cover key applications of time series forecasting—energy, weather, electricity, traffic, and disease—validating the real-world applicability of our model.

\begin{table}[h]
\centering
\caption{Dataset statistics}
\label{table 3}
\vspace*{-2mm}
\resizebox{\columnwidth}{!}{
\begin{tabular}{cccccccc}
\hline
Datasets & ETTh1/h2 & ETTm1/m2 & Weather & Electricity & PEMS08 & ILI & Solar \\ \hline
Variables & 7 & 7 & 21 & 321 & 170 &7 & 137\\ 
Time steps & 17420 & 68680 & 52696 & 26304 & 17856 & 966 & 52560\\ 
Granularity & 1 hour & 15 min & 10 min & 1 hour & 5 min & 1 week & 10 min \\ \hline
\end{tabular}}
\end{table}

\subsection{Baselines Implementation} \label{appendix1.2}
We implemented the baselines based on their official implementations and followed the configurations from their original papers as closely as possible. Due to the lack of GPU memory and time constraint, we reduced the pre-training epochs to 10 and the model size of SimMTM for the Electricity and the PEMS08 datasets. The Transformer encoder in SimMTM is defined with 2 layers, an embedding dimension of 16, a feed-forward network dimension of 16, and 4 heads for the Electricity dataset and 2 layers, an embedding dimension of 8, a feed-forward network dimension of 64, and 4 heads for the PEMS08 dataset. For the PEMS08 dataset, for which other baselines are not implemented in their papers, we followed the default configuration in the official codes. 

For TARNet \cite{tarnet}, which does not evaluate the forecasting performance in the original paper, we implemented TARNet for forecasting. We also implemented Ti-MAE, whose public code is not available. Owing to the large forecast head size of Ti-MAE and our computational resource limits, we could not perform forecasting for the Electricity dataset at prediction lengths of 336 and 720, and for the PEMS08 dataset at a prediction length of 720. Nonetheless, we believe the comparative experiment is valid since Ti-MAE's performance was generally inadequate.

\begin{table}[h]
\caption{Model and pre-training configuration of ST-MTM}
\label{model config}
\vspace{-2mm}
\begin{tabular}{ccc}
\hline
& Hyper-parameters & Candidates \\ \hline
\multirow{4}{*}{Encoder} & Layers & \{1, 2\} \\
&$d_{model}$ & \{16, 32, 64\}  \\
&$n_{head}$ & \{4, 8, 16\} \\
& $d_{ff}$ & \{32, 64, 128\} \\ \hline
Masking & Kernel size & \{25, 50, 100, 200\} \\ \hline
Pre-training & Batch size & \{16, 32, 64, 128\} \\ \hline

\end{tabular}
\end{table}

\subsection{Model and Pre-training Configuration} \label{appendix1.3}
In the pre-training stage, we pre-trained the model with different hyper-parameters according to the datasets. The candidates for the hyper-parameters in the encoder architecture, masking method, and pre-training are summarized in Table \ref{model config}.

For the configuration of the seasonal encoder, we fix the hidden dimension of the per-frequency MLP layer and the MLP layer in the time domain as 128. Other hyper-parameters are fixed as described in the manuscript.

\section{Additional Comparative Evaluation}
\subsection{Performance on Time Series with Minimal Seasonality} \label{appendix2.1}

We have demonstrated the performance of ST-MTM across various seasonal intensities, ranging from ETT which exhibits noisy cyclical patterns, to PEMS08, which displays clear periodic patterns. To fully evaluate its effectiveness under various temporal patterns, it is crucial to assess the robustness of prediction performance when temporal patterns are scarcely discernible, frequently observed in real-world data. For this purpose, previous studies have used the Exchange dataset \cite{autoformer, last}. The Exchange dataset contains daily exchange rates from eight countries from 1990 and to 2016 and is known for minimal discernible periodicity and significant distribution shifts due to the inherent properties of economic data \cite{shao2023exploring}. Since this lack of periodicity poses challenges for forecasting, a model that performs well on the Exchange dataset is considered robust for time series with minimal discernible periodic patterns.

Similarly, we evaluate ST-MTM on the Exchange dataset to demonstrate the robustness of ST-MTM on non-periodic time series forecasting. As shown in Table \ref{table 7}, our model outperforms self-supervised baselines, achieving the best score on six scenarios and the second best score on two scenarios. In addition, our model demonstrates the competitive performance, achieving the second best score on six scenarios compared to decomposition-based forecasting baselines (see Table \ref{table 8}). These results confirm that ST-MTM is robust to time series with weak seasonality and scarce temporal patterns, which prevail in real-world time series. As described in Section 4.4.1, we suggest that this robustness is attributed to the component-wise gating layer, which adaptively determines the interactions between seasonal and trend components to generate effective time series representation.

\subsection{Comparison with TimeSiam}
TimeSiam \cite{timesiam} is the concurrent masked time-series modeling method designed to strengthen temporal modeling capability. TimeSiam extracts relevant temporal information from a past window to supplement the insufficient temporal information in the current masked window and reconstruct it through Siamese networks. While the focuses of ST-MTM and TimeSiam on enhancing temporal modeling capability are distinct, we additionally compare the two methods in time series forecasting. As shown in Table \ref{table timesiam}, ST-MTM outperforms TimeSiam on the ETT datasets and demonstrates competitive performance on other datasets. These results suggest the effectiveness of ST-MTM in modeling temporal dependencies. The results of additional comparative experiments are available on our official repository.

\begin{table*}[t]
\caption{Complete results of multivariate forecasting on the Exchange dataset compared with self-supervised methods in in-domain forecasting scenarios. We fix the input length $L=336$. The best results are in bold and the second best results are underlined. Baselines with * are models adopting a decomposition architecture.}
\label{table 7}
\vspace{-2.5mm}
\resizebox{\textwidth}{!}{%
\begin{tabular}{cc|cccccccccccccccc}
\hline
\multicolumn{2}{c}{Models}  & \multicolumn{2}{c}{ST-MTM} & \multicolumn{2}{c}{SimMTM} & \multicolumn{2}{c}{PatchTST} & \multicolumn{2}{c}{TARNet} & \multicolumn{2}{c}{Ti-MAE}& \multicolumn{2}{c}{TS2Vec} & \multicolumn{2}{c}{CoST *} & \multicolumn{2}{c}{LaST *} \\ \cline{3-18} 
\multicolumn{2}{c}{Metrics} & MSE & MAE & MSE & MAE & MSE & MAE & MSE & MAE & MSE & MAE & MSE & MAE & MSE & MAE & MSE & MAE \\ \hline
\multirow{5}{*}{\rotatebox[origin=c]{90}{Exchange}} & 96 & \textbf{0.093}&	\textbf{0.217}&	0.103&	0.229&	0.107&	0.232&	0.990&	0.849&	0.993&	0.796&	0.466&	0.520&	0.438&	0.501&	\underline{0.096}&	\underline{0.219}\\
 & 192 & \textbf{0.192}&\textbf{0.315}&	0.211&	0.333&	0.216&	0.334&	1.138&	0.914&	1.143&	0.860 &	0.851&	0.700&	0.869&	0.716&	\underline{0.195}&	\underline{0.322}\\
 & 336 & \underline{0.368}&	\underline{0.445}&	0.396&	0.463&	0.422&	0.477&	1.342&	0.981&	2.280&	1.144&	1.444&	0.920&	1.406&	0.913&	\textbf{0.279}&	\textbf{0.395}\\
 & 720 & 1.057&	0.784&	\underline{1.033}&	\underline{0.775}&	\textbf{0.974}&	\textbf{0.735}&	2.961&	1.420&	3.334&	1.471&	1.887&	1.079&	1.902&	1.086&	1.316&	0.846\\ \cline{2-18}
 & Avg & \textbf{0.428} & \textbf{0.440} & 0.436 & 0.450 & \underline{0.430} & \underline{0.444} & 1.608 & 1.041 & 1.937 & 1.068 & 1.162 & 0.805 & 1.154 & 0.804 & 0.471 & 0.445 \\ \hline
\end{tabular}%
}
\end{table*}

\begin{table*}[t]
\caption{Complete results of multivariate forecasting on the Exchange dataset compared with decomposition-based supervised forecasting methods in in-domain forecasting scenarios. We fix the input length $L=336$. The best results are in bold and the second best results are underlined.}
\label{table 8}
\vspace{-2.5mm}
\begin{tabular}{cc|cccccccccccccc}
\hline
\multicolumn{2}{c}{Models}  & \multicolumn{2}{c}{ST-MTM} & \multicolumn{2}{c}{SCNN} & \multicolumn{2}{c}{TimesNet} & \multicolumn{2}{c}{DLinear} & \multicolumn{2}{c}{Autoformer}& \multicolumn{2}{c}{FEDformer} & \multicolumn{2}{c}{ETSformer} \\ \cline{3-16} 
\multicolumn{2}{c}{Metrics} & MSE & MAE & MSE & MAE & MSE & MAE & MSE & MAE & MSE & MAE & MSE & MAE & MSE & MAE \\ \hline
\multirow{5}{*}{\rotatebox[origin=c]{90}{Exchange}} & 96 & \underline{0.093}&	\underline{0.217}&	\textbf{0.089}&	\textbf{0.207}&	0.201&	0.334&	0.110&	0.235&	0.401&	0.487&	0.393&	0.471&	0.097&	0.225\\
 & 192 & \underline{0.192}&	\underline{0.315}&	\textbf{0.182}&	\textbf{0.303}&	0.334&	0.433&	0.263&	0.374&	0.726&	0.661&	0.488&	0.525&	0.194&	0.326 \\
 & 336 & \underline{0.368}&	\underline{0.445}&	\textbf{0.349}&	\textbf{0.428}&	0.571&	0.573&	0.385&	0.470&	0.957&	0.764&	0.690&	0.634&	0.380&	0.449 \\
 & 720 & 1.057&	0.784&	0.968&	\underline{0.735}&	1.664&	0.983&	\textbf{0.761}&	\textbf{0.668}&	1.340&	0.896&	1.464&	0.943&	\underline{0.952}&	0.763 \\ \cline{2-16}
 & Avg & 0.428 & 0.440 & \underline{0.397} & \textbf{0.418} & 0.692 & 0.581 & \textbf{0.380} & \underline{0.437} & 0.856 & 0.702 & 0.759 & 0.643 & 0.406 & 0.441 \\ \hline
\end{tabular}%
\end{table*}

\begin{table}[t]
\caption{Multivariate forecasting results compared with TimeSiam in in-domain forecasting scenarios. The results are averaged from all prediction lengths.}
\label{table timesiam}
\vspace{-2mm}
\begin{tabular}{cc|cccc}
\hline
\multicolumn{2}{c|}{Models}  & \multicolumn{2}{c}{ST-MTM} & \multicolumn{2}{c}{TimeSiam}  \\ \cline{3-6} 
\multicolumn{2}{c|}{Metrics} & MSE & MAE & MSE & MAE  \\ \hline
\multicolumn{2}{c|}{ETTh1} & \textbf{0.413} & \textbf{0.429} & 0.420 & 0.438  \\ 
\multicolumn{2}{c|}{ETTh2} & \textbf{0.344} & \textbf{0.388} & 0.367 & 0.406 \\ 
\multicolumn{2}{c|}{ETTm1} & \textbf{0.350} & \textbf{0.383} & 0.352 & 0.385 \\ 
\multicolumn{2}{c|}{ETTm2} & \textbf{0.253} & \textbf{0.315} & 0.266 & 0.324  \\ 
\multicolumn{2}{c|}{Weather} & 0.230 & 0.276 & \textbf{0.229} & \textbf{0.265} \\ 
\multicolumn{2}{c|}{Electricity} & 0.170 & 0.273 & \textbf{0.159} & \textbf{0.250} \\ 
\multicolumn{2}{c|}{PEMS08} & 0.204 & 0.305 & \textbf{0.187} & \textbf{0.251} \\ 
\multicolumn{2}{c|}{ILI} & 2.757 & \textbf{1.062} & \textbf{2.713} & 1.974  \\ 
\multicolumn{2}{c|}{Solar} & \textbf{0.195} & 0.271 & 0.196 & \textbf{0.248}\\ \hline
\end{tabular}%
\end{table}

\begin{figure}[t]
  \centering
  \includegraphics[width=\linewidth]{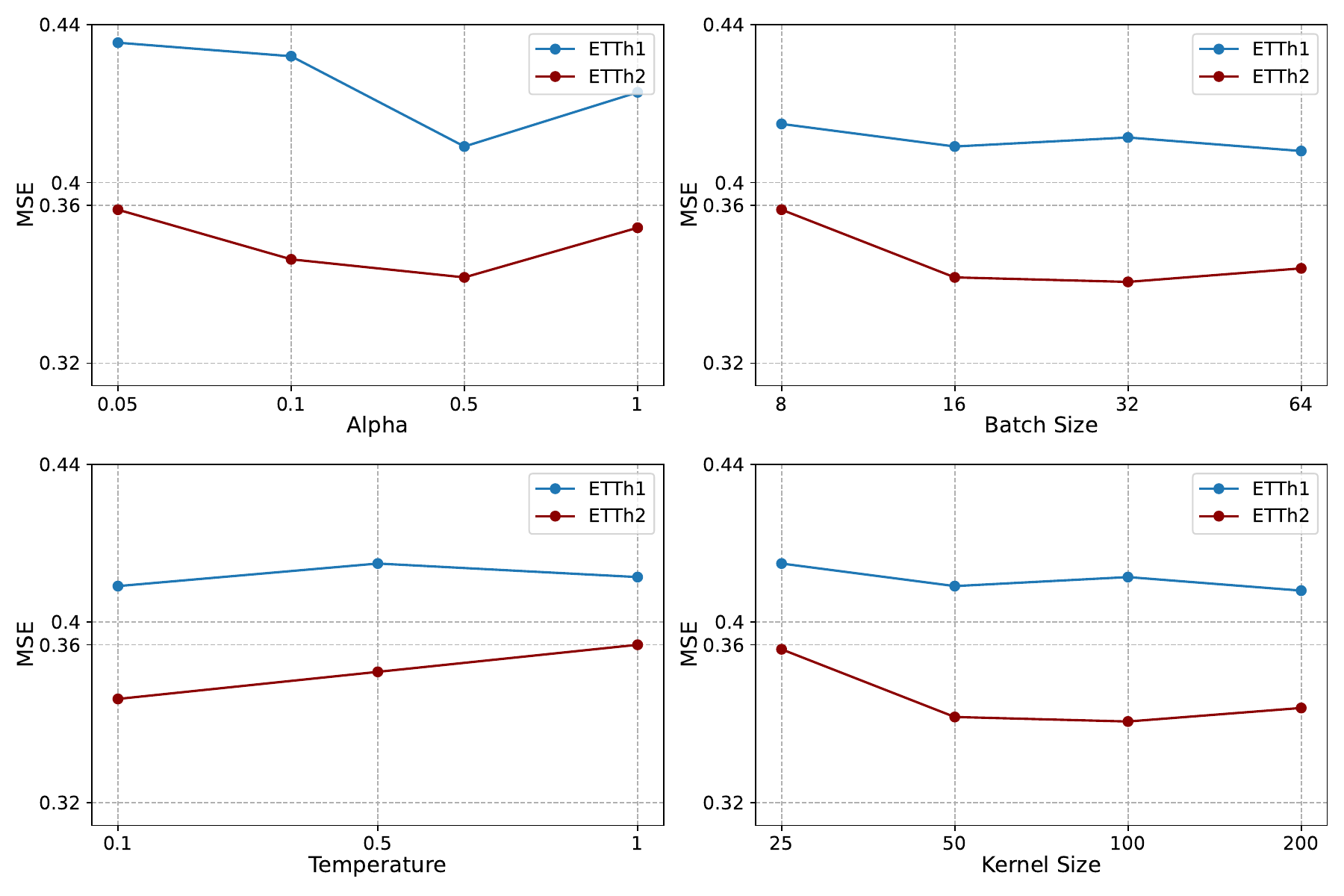}
  \vspace{-5mm}
  \caption{Sensitivity analysis of alpha (upper left), batch size (upper right), temperature (lower left), and kernel size (lower right). We report the average MSE score for all prediction length, \{96, 192, 336, 720\}.}
  \label{figure9}
\end{figure}

\section{Sensitivity analysis} \label{appendix3}
\subsection{Contextual Contrastive Learning} \label{appendix3.1}
We conduct a sensitivity analysis on hyper-parameters for contextual contrastive learning, namely $\alpha$, batch size, and temperature. We experimentally demonstrate that pre-training with contextual contrastive loss enhances the performance of seasonal-trend decomposition in masked time-series modeling. As shown in Figure \ref{figure9}, the regularization parameter $\alpha=0.5$ resulted in the smallest MSE on the ETTh1 and ETTh2 datasets, which is the value used in the main text. The result also indicates that the average MSE gradually decreases as the batch size increases, likely due to the larger number of negative masked seasonal representations available for contextual contrastive loss. We found that contextual contrastive learning benefited from the large batch size. Meanwhile, the performance of our model remained robust across various temperatures.

\begin{figure}[h]
  \centering
  \includegraphics[width=\linewidth]{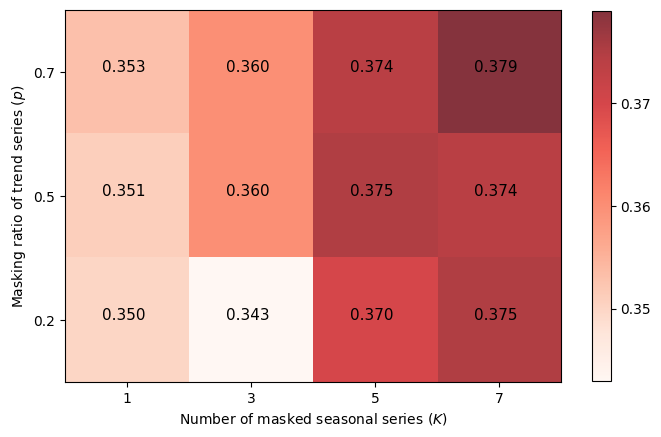}
  \vspace{-5mm}
  \caption{The MSE performance of ST-MTM on the ETTh2 dataset with different masking ratios of trend $p$ and the number of masked seasonal series $K$. We report the average MSE score for all prediction length, \{96, 192, 336, 720\}.}
  \label{figure10}
\end{figure}

\subsection{Moving Average} \label{appendix3.2}
We conduct experiments using various kernel sizes for the moving average operation to extract trends from raw time series. We found that the optimal kernel size varies for each dataset. The best kernel sizes are 200 for ETTh1 and 50 for ETTh2, which approximately correspond to 8 and 2 days, respectively, considering the datasets are at hourly intervals. Note that these optimal kernel sizes are the hyper-parameters with which we report performance. However, we found that the performances are quite robust to the kernel size. Thus, exploratory analysis for extracting a reliable trend should be conducted to decide on the kernel size.

\subsection{Seasonal-Trend Masking}\label{appendix3.3}
Figure \ref{figure10} displays the sensitivity analysis of hyper-parameters for seasonal-trend masking on the ETTh2 dataset. The difficulty of reconstruction increases as the masking ratio is high but decreases as the number of masked series increases. We investigate the relationship between the masked ratio on the trend and the number of masked seasonal series used for reconstruction. The forecasting performance remains robust to variations in the masking ratio of trend series, as indicated by the consistent performance observed vertically in Figure \ref{figure10}. However, ST-MTM shows the higher MSE as the number of masked seasonal series increases, as indicated by the increasing MSE observed horizontally. Given that the difficulty of reconstruction does not directly correlate with forecasting performance, it is challenging to decide a clear tendency for each component. Empirically, we selected a masking ratio of 0.2 and generated three masked seasonal series for pre-training ST-MTM throughout the study.

\begin{table}[t]
\caption{Running time (in seconds) comparison at training phases on the ETTh1 dataset}
\label{runtime}
\vspace{-2mm}
\resizebox{\columnwidth}{!}{
\begin{tabular}{cccccccc}
\hline
Phase & Horizon & ST-MTM & SimMTM&  LaST & SCNN & TimesNet &  ETSformer \\ \hline
Pre-training & - & 450.5 & 625.1 & - &- &- &-\\ \hline
& 96 & 146.0 & 89.3 &182.0 &366.7 &342.0 &288.0\\
Training& 192 &  144.3 & 89.0&189.7 &369.7 &386.3 &295.7\\
(fine-tuning)& 336 & 145.3 & 89.0 &211.3 &368.0 &482.7 &302.0 \\
& 720 & 141.7 & 88.0& 225.3 &376.7 &510.0 &320.7\\ \hline
\end{tabular}
}
\end{table}

\section{Runtime Analysis}\label{appendix4}

Table \ref{runtime} shows the average running time of self-supervised and forecasting methods for each stage on the ETTh1 dataset, measured three times per stage. Pre-training and training epochs are set to 10. For comparison, we include SimMTM from MTM, which uses a vanilla Transformer encoder similar to our model, and LaST from contrastive learning, which incorporates seasonal-trend decomposition. For supervised forecasting baselines, we select SCNN, TimesNet, and ETSformer, as they demonstrate superior performance. All experiments are conducted on a single Nvidia Titan RTX 3080 GPU. The results show that ST-MTM has a shorter pre-training time than SimMTM and a shorter training time than supervised forecasting methods. Additionally, ST-MTM requires only one pre-training step, enabling quick fine-tuning for different forecasting scenarios. This is particularly practical for settings where each prediction horizon would otherwise require a separate forecaster. These findings underscore the utility of ST-MTM in real-world applications.

\section{Forecasting Showcases}\label{appendix5}
\begin{figure}[h]
  \centering
  \includegraphics[width=\linewidth]{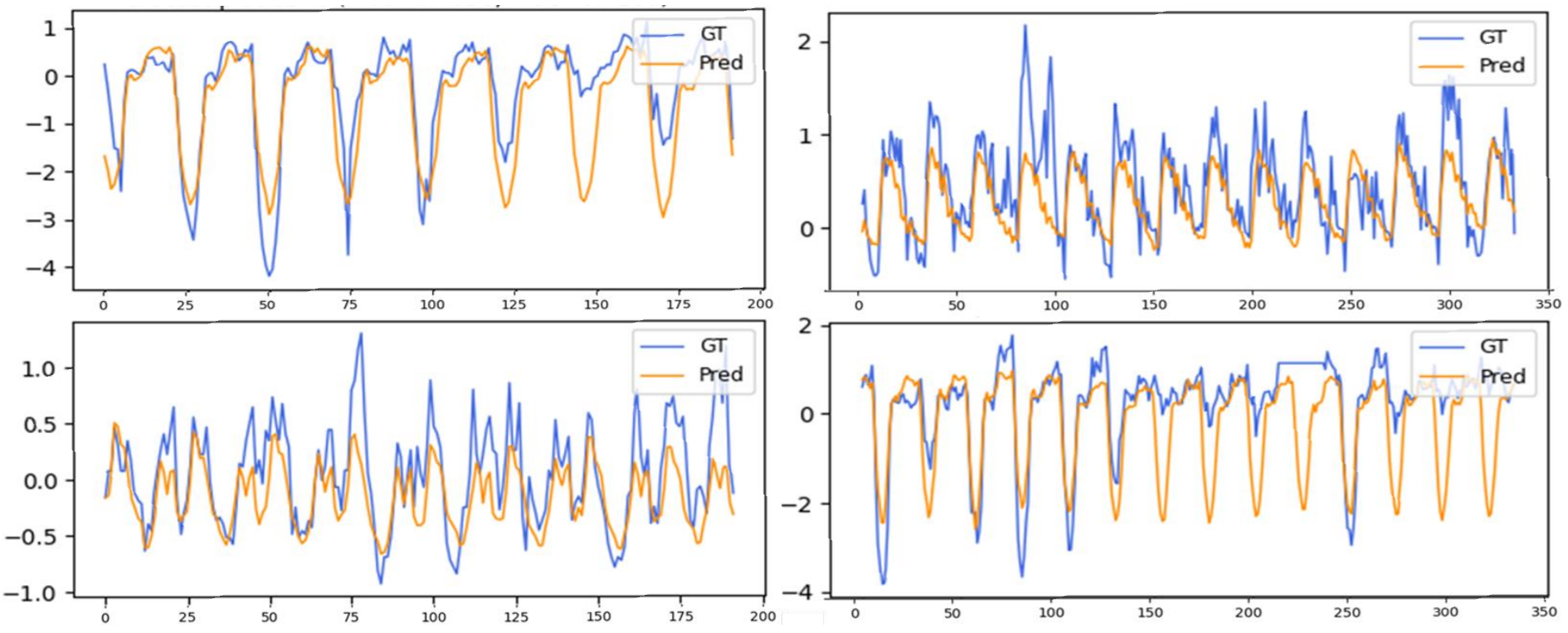}
  \caption{Prediction cases on the ETTh1 dataset for prediction lengths of 192 and 336. The left figures display the prediction of 192 time steps and the right figures display the prediction of 336 time steps. Blue lines represent the ground truth, and yellow lines represent the model predictions.}
  \label{figure ett}
\end{figure}

\begin{figure}[h]
  \centering
  \includegraphics[width=\linewidth]{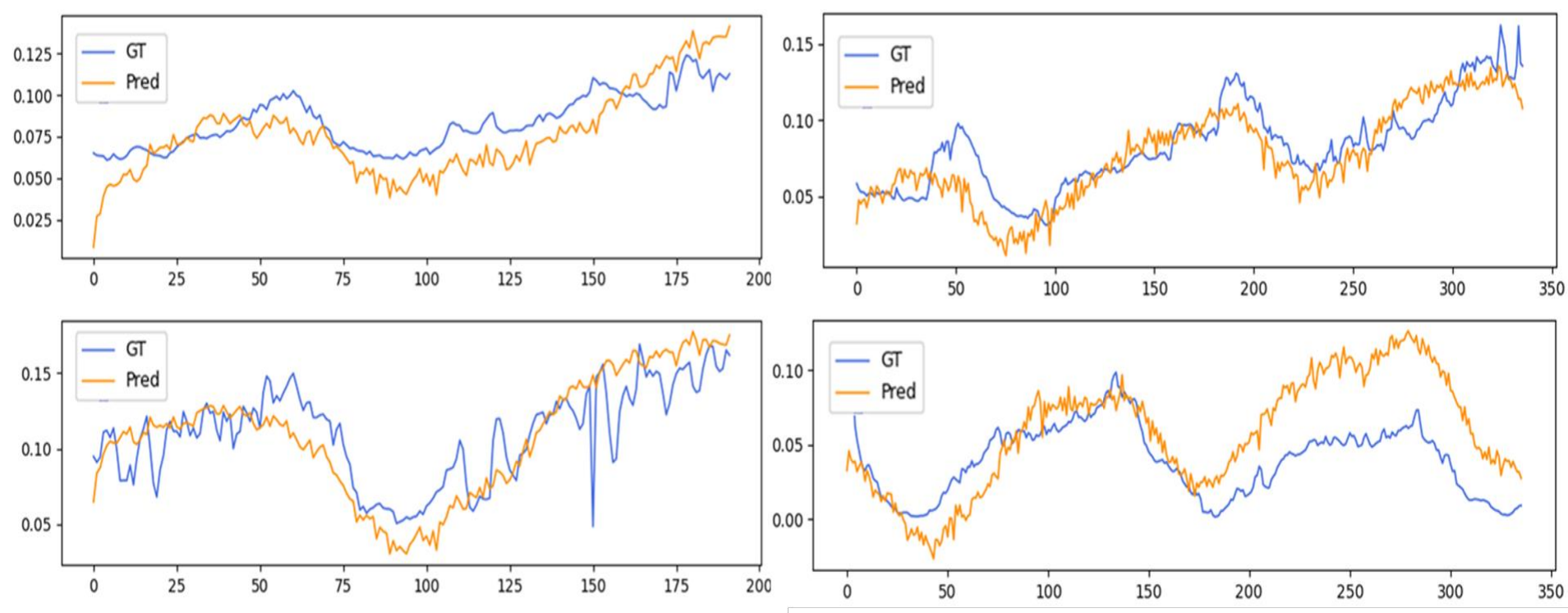}
  \caption{Prediction cases on the Weather dataset for prediction lengths of 192 and 336. The left figures display the prediction of 192 time steps and the right figures display the prediction of 336 time steps.}
  \label{figure weather}
\end{figure}

We visualize the forecasting results of ST-MTM on the ETTh1 and Weather datasets in Figure \ref{figure ett} and \ref{figure weather}.

\section{Acknowledgment}
This research was partly supported by the Basic Science Research Program through the National Research Foundation of Korea (NRF), funded by the Ministry of Education (RS-2024-00413582), as well as by the Institute of Information \& Communications Technology Planning \& Evaluation (IITP) grants funded by the Korean government (MSIT) (RS-2024-00439932, SW Starlab; RS-2020-II201336, Artificial Intelligence Graduate School Program - UNIST; RS-2021-II212068, Artificial Intelligence Innovation Hub; RS-2024-00422098, Global Research Support Program in the Digital Field Program; RS-2024-00443780, Development of Foundation Models for Bioelectrical Signal Data and Validation of Their Clinical Applications: A Noise-and-Variability Robust, Generalizable Self-Supervised Learning Approach).

\end{document}